\documentclass{article}
\usepackage{ijcai15}

\usepackage{times}
\usepackage{epsfig}
\usepackage{epstopdf}
\usepackage{amsmath}
\usepackage{amsfonts}
\usepackage{algorithm}
\usepackage{algorithmic}

\usepackage{float}
\usepackage{multirow}

\title{Reader-Aware Multi-Document Summarization via Sparse Coding\thanks{The work described in this paper is substantially supported by grants
from the Research and Development Grant of Huawei Technologies Co. Ltd (YB2013090068/TH138232) and the Research Grant Council of the Hong Kong Special Administrative Region, China (Project Codes: 413510 and 14203414).}}

\author{Piji Li$^{\dag}$ \ \ Lidong Bing$^{\ddag}$ \ \ Wai Lam$^{\dag}$ \ \ Hang Li$^{\S}$ \and Yi Liao$^{\dag}$\\
$^{\dag}$Department of Systems Engineering and Engineering Management,\\
The Chinese University of Hong Kong, Hong Kong\\
$^{\ddag}$Machine Learning Department, Carnegie Mellon University, Pittsburgh, PA USA\\
$^{\S}$Noah's Ark Lab, Huawei Technologies, Hong Kong\\
$^{\dag}$\{pjli, wlam, yliao\}@se.cuhk.edu.hk, $^{\ddag}$lbing@cs.cmu.edu, $^{\S}$hangli.hl@huawei.com}
\begin{document}

\maketitle

\begin{abstract}
We propose a new MDS paradigm called reader-aware multi-document summarization (RA-MDS).
Specifically, a set of reader comments associated with the news reports are also collected. The generated summaries from the reports for the event should be salient
according to not only the reports but also the reader comments. To tackle this RA-MDS problem, we propose a sparse-coding-based method that is able to calculate the salience of the text units by jointly considering news reports and reader comments. Another reader-aware characteristic of our framework is
to improve linguistic quality via entity rewriting. The rewriting consideration is jointly assessed together with other summarization requirements under a unified optimization model. To support the generation of compressive summaries via optimization, we explore a finer syntactic unit, namely, noun/verb phrase. In this work, we also generate a data set for conducting RA-MDS. Extensive experiments on this data set and some classical data sets demonstrate the effectiveness of our proposed approach.
\end{abstract}

\section{Introduction}

In the typical multi-document summarization (MDS) setting, the input is a set of documents/reports about
the same topic/event. The reports on the same event normally
cover many aspects and the continuous follow-up reports bring in more information of it.
Therefore, it is very challenging to generate a short and salient summary for an event. MDS has drawn some attention and some method have been proposed.
For example, Wan et al.~\shortcite{Wan:2007:MBT:1625275.1625743} proposed an extraction-based approach that employs a manifold ranking method to calculate the salience of each sentence. Filatova and Hatzivassiloglou~\shortcite{Filatova:2004:FMI:1220355.1220412}
modeled the MDS task as an instance of the maximum coverage
set problem. Gillick and Favre~\shortcite{Gillick:2009:SGM:1611638.1611640}
developed an exact solution for a model similar to~\cite{Filatova:2004:FMI:1220355.1220412} based on the weighted sum of the concepts (approximated by bigrams). \cite{li2013document} proposed a guided sentence compression framework to generate compressive summaries by training a conditional random field (CRF) based on a annotated corpus. \cite{liimproving} considered linguistic quality in their framework. \cite{ngexploiting} exploited timelines to enhance MDS. Moreover, many works \cite{liu2012query,kaageback2014extractive,denil2014modelling,cao2015ranking} utilized deep learning techniques to tackle summarization tasks.

As more and more user generated content is available,
one natural extension of the setting is to incorporate such content
regarding the event so as to directly or indirectly improve the generated summaries with greater user satisfaction.
In this paper, we investigate
a new setting in this direction. Specifically, a set of reader comments
associated with the news reports are also collected. The generated
summaries from the reports for the event should be salient
according to not only the reports but also the reader comments.
We name such a paradigm of extension as reader-aware multi-document
summarization (RA-MDS).

We give a real example taken from a data set collected by us to illustrate the
importance of RA-MDS. One hot event in 2014 is ``Malaysia Airlines jet MH370 disappeared''.
After the outbreak of this event, lots of reports are posted on different news media.
Most existing summarization systems can only create summaries with general information,
e.g., ``Flight MH370, carrying 227 passengers and 12 crew members, vanished early Saturday after departing Kuala Lumpur for Beijing'',
due to the fact that they extract information solely from the report content.
However, after analyzing the reader comments, we find that many readers are interested in more specific aspects,
such as ``Military radar indicated that the plane may have turned from its flight route before losing contact'' and
``Two passengers who appear to have used stolen European passports to board''.
Under the RA-MDS setting, one should jointly consider news and comments when generating the summary
so that the summary content can cover not only important aspects of the event,
but also aspects that attract reader interests as reflected in the reader comments.


No previous work has
investigated how to incorporate the comments in MDS problem.
One challenge is how to conduct salience
calculation by jointly considering the focus of news reports and the reader interests revealed by comments. Meanwhile, the model should not be sensitive to the availability of diverse aspects of reader comments.
Another challenge is that reader comments are very noisy, grammatically and
informatively. Some previous works explore the effect of comments or social contexts in
single document summarization (such as blog summarization)~\cite{Hu:2008:CDS:1390334.1390385,Yang:2011:SCS:2009916.2009954}. However, the problem setting of RA-MDS is more challenging because the considered
comments are about an event with multiple reports spanning a time period,
resulting in diverse and noisy comments.

To tackle the above challenges, we propose a sparse-coding-based
method that is able to calculate the salience of the text units
by jointly considering news reports and reader comments.
Intuitively, the nature of summarization is to select a small number of
semantic units to reconstruct the original semantic space of the whole topic.
In our RA-MDS setting, the semantic space incorporates both the news and
reader comments.
The selected semantic units are sparse and hold the semantic diversity property.
Then one issue is how to find these sparse and diverse
semantic units efficiently without supervised training data.
Sparse coding is a suitable method for learning sets of over-complete bases to
represent data efficiently, and it has been demonstrated to be very useful in computer vision \cite{mairal2014sparse}. Moreover, sparse coding can jointly
consider news and comments to select semantic units in a very simple
and elegant way, by just adding a comments reconstruction error item into the
original loss function. Currently, there are only a few works employing sparse
coding for the summarization task. DSDR \cite{he2012document} represents each
sentence as a non-negative linear combination of summary sentences. But this method
does not consider the sparsity. MDS-Sparse \cite{liu-mds-sparse} proposed a two-level
sparse representation model, considering coverage, sparsity, and diversity. But
their results do not show a significant improvement.
In this paper, we propose a more efficient and direct sparse model to tackle these problems and achieve encouraging results on different data sets.

Another reader-aware characteristic of our framework is
to improve linguistic quality via entity rewriting.
Summaries may contain phrases that are not understandable out
of context since the sentences compiled from different documents might contain
too little, too much, or repeated information about the referent.
A human summary writer only uses the full-form mention (e.g. President Barack
Obama) of an entity one time and uses the short-form mention (e.g. Obama)
in the other places. Analogously, for a particular entity, our framework requires that the full-form mention
of the entity should only appear one time in the summary and its other appearances should use the most concise form.
Some early works perform rewriting along with
the greedy selection of individual sentence~\cite{DBLP:conf/ijcnlp/Nenkova08}.
Some other works perform summary rewriting as a post-processing step~\cite{Siddharthan:2011:ISD:2077692.2077699}.
In contrast with such works, the rewriting consideration in our framework is jointly assessed together with other summarization requirements under a unified optimization model.
This brings in two advantages.
First, the assessment of rewriting operation is jointly considered with the generation
of the compressive summary so that it has a global view to generate better
rewriting results. Second, we can make full use of the length limit
because the effect of rewriting operation on summary length is simultaneously
considered with other constraints in the model.
To support the generation of compressive summaries via optimization, we explore a finer syntactic unit, namely, noun/verb phrase.
Precisely, we first decompose the sentences into noun/verb phrases and the salience of
each phrase is calculated by jointly considering its importance in reports and comments.

In this work, we also generate a data set for conducting RA-MDS.
Extensive experiments on our data set and some benchmark data sets have been conducted to examine the efficacy of our framework.

\begin{figure}
\centering
\epsfig{file=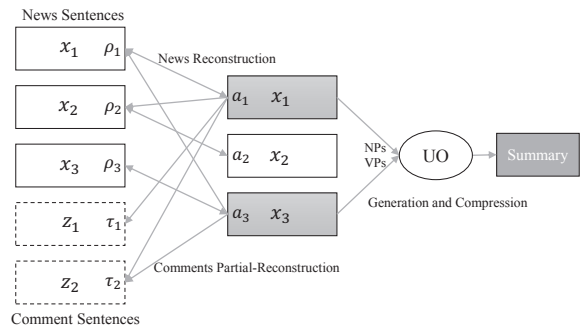, width=3in}
\caption{Our RA-MDS framework.
\vspace{-0.3cm}}
\label{fig:framework}
\end{figure}

\section{Description of the Proposed Framework}

\subsection{Overview}
To tackle the RA-MDS problem, we propose an unsupervised compressive summarization framework.
The overview of our framework is depicted in Fig.~\ref{fig:framework}.
A sparse-coding-based method is proposed to reconstruct
the semantic space of a topic, revealed by both the news sentences
i.e., $x_i$'s and the
comment sentences i.e., $z_i$'s, on the news sentences. Thus, an expressiveness score $a_i$ is designed for each news sentence.
The dashed boxes of comment sentences indicate that a special treatment is applied on comments to avoid noise in the reconstruction.
The details will be introduced in Section~\ref{sec:sent_expres}.
The compression is carried out by deleting the unimportant constituents,
i.e. phrases, of the input sentence. We first decompose each sentence into noun
phrases (NPs) and verb phrases (VPs).
The salience of a phrase depends on two criteria,
namely, the expressiveness score inherited from the sentence, and the concept score of the phrase.
The extraction of phrases and the calculation of phrase salience will be
introduced in Section~\ref{sec:np_vp_salience}.
Our framework carries out mention rewriting for entities to improve
the linguistic quality of our summary.
Specifically, we rewrite the mentions of three types of named entities, namely,
person, location, and organization. We will discuss the details of mention detection,
mention cluster merging, short-form and full-form mention finding in
Section~\ref{sec:mention_prepare}.
After the above preparation steps, we will introduce our
summarization model in Section~\ref{sec:ILP_framework}. Our model
simultaneously performs sentence compression and mention rewriting
via a unified optimization method. Meanwhile,
a variety of summarization requirements are considered
via formulating them as the constraints.



\subsection{\mbox{Reader-Aware Sentence Expressiveness}}
\label{sec:sent_expres}

Intuitively, the nature of summarization is to select semantic units which can be used to reconstruct the original semantic space of the topic.
The expressiveness score of a sentence in the news is defined as its
contribution in constructing the semantic space of the topic from both the news content and the
reader comments. Therefore, the expressiveness
conveys the attention that a sentence attracts
from both the news writers and the readers.
We propose a sparse coding model to compute such expressiveness scores.

In typical sparse coding, the aim is to find a set of
basis vectors $\mathbf{\phi}_i$ which can be used to reconstruct
$m$ target/input vectors $\mathbf{x}_i$ as a linear combination of them
so as to minimize the following loss function:
\begin{equation}
\mathop {{\rm{min}}}\limits_{A,\Psi } \sum\limits_{i = 1}^m {{ \| {\mathbf{x}_i - \sum\limits_{j = 1}^k {{a_{ij}}} {\mathbf{\phi} _j}} \|_2^2}}  + \lambda \sum\limits_{j = 1}^k S ({a_{ij}})
\end{equation}
where $S(.)$ is a sparsity cost function which penalizes $a_i$ for being far from zero.

In our summarization task, each topic contains a set of news reports and a set of reader comments.
After stemming and stop-word removal, we build a
dictionary for the topic by using unigrams and bigrams from the
news. Then, each sentence of news and comments is represented as
a weighted term-frequency vector. Let $\mathbf{X} = \{\mathbf{x}_1, \mathbf{x}_2,\ldots, \mathbf{x}_m\}$
and $\mathbf{Z} = \{\mathbf{z}_1, \mathbf{z}_2,\ldots, \mathbf{z}_n\}$ denote the vectors of sentences
from news and comments respectively, where $\mathbf{x}_i \in\mathbb{R}^d$
and $\mathbf{z}_i \in \mathbb{R}^d$ are term-frequency vectors.
There are $d$ terms in dictionary, $m$ sentences
in news, and $n$ sentences in comments for each topic.
We take semantic units as sentences here, and assume that for each sentence $\mathbf{x}_i$, there is a coefficient variable $a_i$, named expressiveness
score, to represent the contribution of this sentence in the reconstruction.

Based on the spirit of sparse coding, we directly regard each news sentence
$\mathbf{x}_i$ as a candidate basis vector, and all $\mathbf{x}_i$'s are employed to
reconstruct the semantic space of the topic, including $\mathbf{X}$ and $\mathbf{Z}$.
Thus we propose a preliminary error formulation as expressed in Eq.~\ref{e:pre_error} for which we aim at minimizing:
\begin{equation}
\label{e:pre_error}
\frac{1}{{2m}}\sum\limits_{i = 1}^m {\| {{\mathbf{x}_i} - \sum\limits_{j = 1}^m {{a_j}{\mathbf{x}_j}} } \|_2^2} + \frac{1}{{2n}}\sum\limits_{i = 1}^n {\| {{\mathbf{z}_i} - \sum\limits_{j = 1}^m {{a_j}{\mathbf{x}_j}} } \|_2^2}
\end{equation}
where the coefficient $a_j$'s are the expressiveness scores and all the target vectors share the same coefficient vector $A$ here.

To harness the characteristics of the summarization problem setting more effectively, we refine
the preliminary error formulation as given in Eq.~\ref{e:pre_error} along three directions.
(1) As mentioned before, the original sentence vector space can be
constructed by a subset of them, i.e., the number of summary
sentences is sparse, so we put a sparsity constraint on the
coefficient vector $A$ using $L_1$-norm $\lambda {\| A \|_1}$ in Eq.~\ref{e:pre_error}, with the weight $\lambda$ as a scaling
constant to determine its relative importance. Moreover, we just consider non-negative linear reconstruction in our framework, so we add non-negative constraints on the coefficients.
(2) As previous work \cite{Ng_swing:exploiting} mentioned, some prior knowledge can benefit the sentence expressiveness detection performance, e.g., sentence position. So we add a variable $\rho_i$ to weight each news-sentence reconstruction error. Here, we employ the position information to generate $\rho$:
\begin{equation}
\rho  = \left\{ \begin{array}{l}
{C^p}, \ \ if \  p < \overline p \\
{C^{\overline p }}, \ otherwise
\end{array} \right.
\end{equation}
where $p$ is the paragraph ID in each document starting from $0$, and $C$ is a positive constant which smaller than 1.
(3) Besides those useful information, comments usually introduce lots of noise data. To tackle this problem, our first step is to eliminate terms only appear in comments; another step is to add a parameter $\tau_i$ to control the comment-sentence reconstruction error. Due to the fact that the semantic units of generated summaries are all from news, intuitively, a comment-sentence will introduce more information if it is more similar with news. Therefore, we employ the mean cosine similarity between comment-sentence $\mathbf{z}_i$ with all the news-sentences $\mathbf{X}$ as the weight variable $\tau_i$.

After the above considerations, we have the global loss function as follows:
\begin{small}
\begin{equation}
\label{e:loss}
\begin{aligned}
J = &\mathop {\min }\limits_A \frac{1}{{2m}}\sum\limits_{i = 1}^m {\rho_i}{\| {{\mathbf{x}_i} - \sum\limits_{j = 1}^m {{a_j}{\mathbf{x}_j}} } \|_2^2} \\
&+ \frac{1}{{2n}}\sum\limits_{i = 1}^n {\tau_i}{\| {{\mathbf{z}_i} - \sum\limits_{j = 1}^m {{a_j}{\mathbf{x}_j}} } \|_2^2} + \lambda {\| A \|_1} \\
&s.t. \ \ \ a_j \ge 0 \ \text{for} \ j \in \{1,...,m\}, \lambda > 0
\end{aligned}
\end{equation}
\end{small}

\begin{small}
\begin{algorithm}[!t]
  \caption{Coordinate descent algorithm for sentence expressiveness detection}
  \label{alg:sc}
  \begin{algorithmic}[1]
  \REQUIRE News sentences $\mathbf{X} \in \mathbb{R}^{d \times m}$, comments sentences $\mathbf{Z} \in \mathbb{R}^{d \times n}$, news reconstruction weight $\rho_i$, comments reconstruction weight $\tau_i$, penalty parameter $\lambda$, and stopping criterion $T$ and $\varepsilon$.
  \ENSURE Salience vector $A^* \in \mathbb{R}^m$.
  \STATE Initialize $A \leftarrow \textbf{0}$, $t \leftarrow 0$;
  \WHILE{$t < T$ and $J_\varepsilon ^t > \varepsilon$}
  \STATE reconstructing:
        $\mathbf{\bar x} = \sum\limits_{j = 1}^m {a_j^t{\mathbf{x}_j}}$
  \STATE take partial derivatives for reconstruction error items:
    \begin{equation}
      \begin{aligned}
        \frac{{\partial J}}{{\partial {a_k}}} = &- \frac{1}{m}\sum\limits_{i = 1}^m {\rho _i}{{({\mathbf{x}_i} - \mathbf{\bar x})}^ \mathrm{T}}{\mathbf{x}_k}\\
        &- \frac{1}{n}\sum\limits_{i = 1}^n {{\tau _i}{{({\mathbf{z}_i} - \mathbf{\bar x})}^ \mathrm{T}}{\mathbf{x}_k}}
      \end{aligned}
    \end{equation}
  \STATE select the coordinate with maximum partial derivative:\\
    \begin{equation}
        \hat k = \mathop {\arg \max }\limits_{k = 1 \ldots m} \left| {\frac{{\partial J}}{{\partial {a_k}}}} \right|
    \end{equation}
  \STATE update the coordinate by soft-thresholding \cite{donoho1994ideal}:\\
    \begin{equation}
        a_{\hat k}^{t + 1} \leftarrow {S_\lambda }(a_{\hat k}^t - \eta \frac{{\partial J}}{{\partial {a_{\hat k}}}})
        \label{eq:new_a}
    \end{equation}
    where ${S_\lambda }:{a_i} \mapsto sign({a_i})max(\left| {{a_i}} \right| - \lambda ,0)$.
  \STATE ${J_\varepsilon ^t} \leftarrow {J_{{A^{t + 1}}}} - {J_{{A^{t}}}}$, $t \leftarrow t + 1$
  \ENDWHILE
  \RETURN $A^* = A$.
  \end{algorithmic}
\end{algorithm}
\end{small}

For the optimization problem of sparse coding, there are already many classical algorithms \cite{mairal2014sparse}.
In this paper, we utilize Coordinate Descent method as shown in Algorithm \ref{alg:sc}. Under the iterative updating rule as in Eq.~\ref{eq:new_a}, the objective function $J$ is non-increasing, and the convergence of the iteration is guaranteed.

Our sparse coding model introduces several advantages.
First, sparse coding is a class of unsupervised methods, so no manual annotations for training data are needed.
Second, the optimization procedure is modular leading to easily plug in different loss functions.
Third, our model incorporates semantic diversity naturally, as mentioned in \cite{he2012document}.
Last but not the least, it helps the subsequent unified optimization component which generates compressive summaries. In particular, it reduces the number of variables because the sparsity constraint can generate sparse expressiveness scores, i.e., most of the sentences get a $0$ score.

\subsection{Phrase Extraction and Salience Calculation}
\label{sec:np_vp_salience}
\begin{figure}[!t]
\centering
\includegraphics[width=3.35in, height=1.6in]{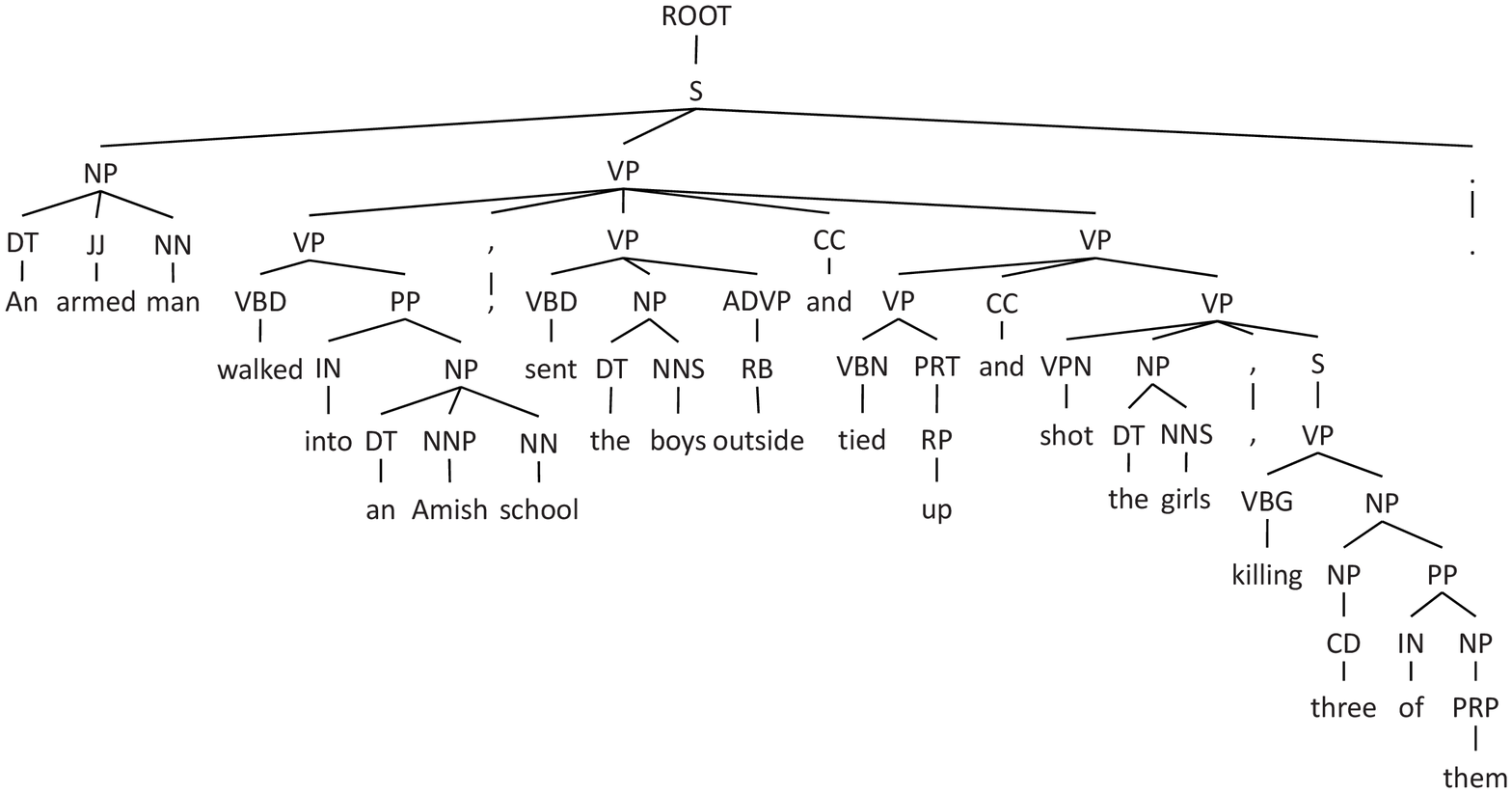}
\caption{\label{fig:syntactic_tree}The constituency tree of a sentence. }
\vspace{-0.3cm}
\end{figure}

We employ Stanford parser \cite{Klein:2003:AUP:1075096.1075150}
to obtain a constituency tree for each input sentence. After that,
we extract NPs and VPs from the tree as follows: (1) The NPs and VPs that are the direct
children of the \textbf{S} node are extracted.
(2) VPs (NPs) in a path on which all the nodes are all VPs (NPs) are also recursively extracted
and regarded as having the same parent node \textbf{S}.
Recursive operation in the second step will only be carried out in two levels
since the phrases in the lower levels may not be able to convey a complete fact.
Take the tree in Fig.~\ref{fig:syntactic_tree} as an example, the corresponding sentence is decomposed into
phrases ``An armed man'', ``walked into an Amish school, sent the boys outside and tied up and shot the girls, killing three of them'', ``walked into an Amish school'', ``sent the boys outside'', and ``tied up and shot the girls, killing three of them''. \footnote{
Because of the recursive operation, the extracted phrases may have overlaps.
Later, we will show how to avoid such overlapping in phrase extraction.
We only consider the recursive operation for a VP with more than one
parallel sub-VPs, such as the highest VP in Fig.~\ref{fig:syntactic_tree}.
The sub-VPs following modal, link or auxiliary verbs are not extracted
as individual VPs. In addition, we also extract the clauses functioning as subjects of
sentences as NPs, such as ``that clause''. Note that we also mention such clauses
as ``noun phrase'' although their labels in the tree could be ``SBAR'' or ``S''.}

The salience of a phrase depends on two criteria.
The first criterion is the expressiveness score which is inherited from the corresponding sentence in the output of our sparse coding model.
The second criterion is the concept score that conveys
the overall importance of the individual concepts in the phrase.
Let $tf(t)$ be the frequency of the term $t$ (unigram/bigram) in the whole topic.
The salience $S_i$ of the phrase $P_i$ is defined as:
\begin{equation}
\label{equ:si}
{S_i} = \frac{{\sum\limits_{t \in P_i} {tf(t)} }}{{\sum\limits_{t \in Topic} {tf(t)} }} \times {a_i},
\end{equation}
where $a_i$ is the expressiveness of the sentence containing $P_i$.

\subsection{Preparation of Entity Mentions for Rewriting}
\label{sec:mention_prepare}

We first conduct co-reference resolution
for each document using Stanford co-reference resolution package~\cite{Lee:2013:DCR:2576217.2576221}.
We adopt those resolution rules that are able to achieve high quality
and address our need for summarization. In particular,
Sieve 1, 2, 3, 4, 5, 9, and 10 in the package are employed.
A set of clusters are obtained and each cluster contains
the mentions corresponding to the same entity in a document.
The clusters from different documents in the same topic are merged
by matching the named entities. Three types of entities are considered, namely,
person, location, and organization.

Let $M$ denote the mention cluster of an entity.
The full-form mention $m^f$ is determined as:
\begin{equation}
{m^f} = \mathop {\arg \max }\limits_{m \in M} \sum\limits_{t \in m} {tf'(t)}
\end{equation}
where $tf'(t)$ is calculated in $M$.
We do not simply select the longest one since it could be too verbose.
The short-form mention $m^s$ is determined as:
\begin{equation}
{m^s} = \mathop {\arg \max }\limits_{m \in M'} \sum\limits_{t \in m} {tf'(t)}
\end{equation}
where $M'$ contains the mentions that are the shortest and meanwhile are not
pronouns.

\subsection{Unified Optimization Framework}
\label{sec:ILP_framework}

The objective function of our optimization formulation is defined as:
\begin{small}
\begin{equation}
\label{e:objective}
\max\{\sum_i{\alpha_i S_i} - \sum_{i<j}{\alpha_{ij}(S_i+S_j)R_{ij}}\},
\end{equation}
\end{small}
where $\alpha_i$ is the selection indicator for the phrase $P_i$,
$S_i$ is the salience scores of $P_i$, $\alpha_{ij}$  and $R_{ij}$ is co-occurrence indicator
and the similarity a pair of phrases ($P_i$, $P_j$) respectively.
The similarity is calculated with the Jaccard Index based method.
Specifically, this objective maximizes the salience score of the selected phrases
as indicated by the first term, and penalizes the selection of similar phrase pairs.
The constraints that govern the selected phrases are able to form compressive sentences and the constraints for entity rewriting
are given below. Note that the rewriting consideration is conducted for different candidates for the purpose of the assessment of the effects on summarization in the optimization framework. Consequently, no actual permanent rewriting operations are conducted during the optimization process. The actual rewriting operations will be carried out on the selected phrases output from the optimization component in the post-processing stage.

\noindent\textbf{\texttt{Compressive sentence generation}}.
Let $\beta_k$ denote the selection indicator of sentence $x_k$. If any phrase from $x_k$
is selected, $\beta_k=1$. Otherwise, $\beta_k=0$.
For generating a compressed summary sentence, it is required that if
$\beta_k=1$, at least one NP
and at lease one VP of the sentence should be selected. It is expressed as:
\begin{equation}
\forall{P_i\in x_k \wedge P_i~is~an~NP}, \alpha_i \le \beta_k \wedge \sum_{i}{\alpha_i} \ge \beta_k,
\end{equation}
\begin{equation}
\forall{P_i\in x_k \wedge P_i~is~a~VP}, \alpha_i \le \beta_k \wedge \sum_{i}{\alpha_i} \ge \beta_k.
\end{equation}

\noindent\textbf{\texttt{Entity rewriting}}.
Let $\mathbf{P}_M$ be the phrases that contain the entity
corresponding to the cluster $M$. For each $P_i\in \mathbf{P}_M$,
two indicators $\gamma_i^f$ and $\gamma_i^s$ are defined. $\gamma_i^f$
indicates that the entity in $P_i$ is rewritten by the
full-form, while $\gamma_i^s$ indicates that the entity in $P_i$ is
rewritten by the short-form. To adopt our rewriting strategy, we design the following constraints:
\begin{equation}
\mbox{if } \exists P_i\in \mathbf{P}_M \wedge \alpha_i=1, \sum\limits_{P_j\in \mathbf{P}_M} {\gamma_j^f=1},
\label{eq:ew1}
\end{equation}
\begin{equation}
\mbox{if } P_i\in \mathbf{P}_M \wedge \alpha_i=1, \gamma_i^f+\gamma_i^s=1.
\label{eq:ew2}
\end{equation}

Note that if a phrase contains several mentions of the same entity, we can safely
rewrite the latter appearances with the short-form mention and we only need to decide the rewriting strategy for the first appearance.



\noindent\textbf{\texttt{Not i-within-i}}.
Two phrases in the same path of the constituency tree cannot
be selected at the same time:
\begin{small}
  \begin{equation}\label{}
\mbox{if } \exists{P_k\rightsquigarrow P_j}, \mbox{then } \alpha_k+\alpha_j \leq 1,
\end{equation}
\end{small}
For example, ``walked into an Amish school, sent the boys outside and tied up and shot the girls, killing three of them'' and ``walked into an Amish school'' cannot be both selected.

\noindent\textbf{\texttt{Phrase co-occurrence}}.
These constraints control the co-occurrence relation of two phrases:
\begin{small}
\begin{equation}\label{e:co-occurrence_np}
     \alpha_{ij}-\alpha_{i}\leq 0,~~\alpha_{ij}-\alpha_{j}\leq 0,~~\alpha_{i}+\alpha_{j}-\alpha_{ij}\leq 1;
\end{equation}
\end{small}
The first two constraints state that if
the summary includes both the units $P_i$ and $P_j$, then we have to include them individually. The third constraint is the inverse of the first two.

\noindent\textbf{\texttt{Short sentence avoidance}}.
We do not select the VPs from the sentences shorter than
a threshold because a short sentence normally cannot convey a complete key fact

\noindent\textbf{\texttt{Pronoun avoidance}}.
As previously observed \cite{Woodsend:2012:MAS:2390948.2390978},
 pronouns are normally not used by human summary writers.
We exclude the NPs that are pronouns from being selected.

\noindent\textbf{\texttt{Length constraint}}. The overall length of the selected NPs and VPs is no larger than a limit $L$. Note that the length
calculation considers the effect of rewriting operations via the
rewriting indicators.

The objective function and constraints are linear
so that the optimization can be solved by existing Integer Linear Programming (ILP)
solvers such as simplex algorithm~\cite{Dantzig:1997:LPI:248375}. In the implementation, we use a package called lp\_solve\footnote{http://lpsolve.sourceforge.net/5.5/}.


\subsection{Postprocessing}

The timestamp of a summary sentence is defined as the timestamp of the
corresponding document.
The sentences are ordered based on their pseudo-timestamps.
The sentences from the same document are ordered according to
their original order.
Finally, we conduct the appropriate entity rewriting as indicated from the optimization output.

\section{Experiments}
\subsection{Experimental Setting}


\textbf{Our data set}.
Our data set contains 37 topics. Each topic contains 10 related news reports and at least 200 reader comments.
For each topic, we employ summary writers with journalist background to write model summaries. When writing summaries, they take into account the interest of readers
by digesting the reader comments of the event. 3 model summaries are written for each topic. We also have a separate development (tuning) set containing 24 topics and each topic has one model summary.

\textbf{DUC}.
In order to show that our sparse coding based framework can also work well on traditional MDS task, we employ the benchmark data sets DUC 2006 and DUC 2007 for evaluation.
DUC 2006 and DUC 2007 contain 50 and 45 topics respectively. Each topic has 25 news documents and 4 model summaries. The length of the model summary is limited by 250 words.


\textbf{Evaluation metric}.
We use ROUGE score as our evaluation metric \cite{lin2004rouge}\footnote{http://www.berouge.com}
and the F-measures of ROUGE-1, ROUGE-2 and ROUGE-SU4 are reported.

\textbf{Parameter settings}.
We set $C = 0.8$ and $\overline p = 4$ in the position weight function.
For the sparse coding model, we set the stopping criteria $T = 300$, $\varepsilon = 10^{-4}$, and the learning rate $\eta = 1$. For the sparsity item penalty, we set $\lambda = 0.005$.



\subsection{Results on Our Data Set}
\label{sec:result_our_dateset}
We compare our system with three summarization baselines.
\textbf{Random} baseline selects sentences randomly for each topic.
\textbf{Lead} baseline~\cite{wasson1998using} ranks the news chronologically and extracts the leading sentences one by one.
\textbf{MEAD} \cite{radev2004mead}\footnote{http://www.summarization.com/mead/} generates summaries using cluster centroids produced by a topic detection and tracking system.

\begin{table}[!ht]
\centering
\begin{tabular}{p{2.5cm} c c c}
\hline
\textbf{System}  & \textbf{Rouge-1} & \textbf{Rouge-2} & \textbf{Rouge-SU4} \\
\hline
Random & 0.334 & 0.069 & 0.109 \\
Lead & 0.355 & 0.098 & 0.133 \\
MEAD & 0.406 & 0.127 & 0.161 \\
Ours  & \textbf{0.438} & \textbf{0.155} & \textbf{0.186} \\
\hline
\end{tabular}
\caption{Results on our data set.}
\label{tab:our}
\end{table}

As shown in Table \ref{tab:our}, our system reports the best results on all of ROUGE metrics. The reasons are as follows:
(1) Our sparse coding model directly assigns coefficient values as expressiveness scores to the news sentences, which are obtained by minimizing the global semantic space reconstruction error and are able to precisely represent the importance of sentences.
(2) The model can jointly consider news content and reader comments taking into account of more reader-aware information.
(3) In our sparse coding model, we weight the reconstruction error by a prior knowledge, i.e., paragraph position, which can improve the summarization performance significantly.
(4) Our unified optimization framework can further filter the unimportant NPs and VPs and generate the compressed summaries.
(5) We conduct entity rewriting in the unified optimization framework in order to improve the linguistic quality.

\subsection{Results on DUC}
In order to illustrate the performance of our framework on traditional MDS task, we compare it with several state-of-the-art systems on standard data set DUC. Our framework can still be used for MDS task without reader comments by ignoring those components for comments.

Besides Random and Lead methods, we compare our system with two other unsupervised sparse coding based methods, namely DSDR \cite{he2012document} and MDS-Sparse \cite{liu-mds-sparse} (MDS-Sparse+div and MDS-Sparse-div). Because both data set and evaluation metrics are standard, we directly retrieve the results in their papers.
The results are given in Tables~\ref{tab:duc2006} and \ref{tab:duc2007}.
Our system can significantly outperform the comparison methods for the reasons mentioned in Section \ref{sec:result_our_dateset}.

\begin{table}[H]
\centering
\begin{tabular}{p{2.5cm} c c c}
\hline
\textbf{System}  & \textbf{Rouge-1} & \textbf{Rouge-2} & \textbf{Rouge-SU4} \\
\hline
Random & 0.280 & 0.046 & 0.088 \\
Lead & 0.308 & 0.048 & 0.087 \\
DSDR-non & 0.332 & 0.060 & - \\
MDS-Sparse+div       & 0.340 & 0.052 & 0.107 \\
MDS-Sparse-div       & 0.344 & 0.051 & 0.107 \\
Ours       & \textbf{0.391} & \textbf{0.081} & \textbf{0.136} \\
\hline
\end{tabular}
\caption{Results on DUC 2006.}
\label{tab:duc2006}
\end{table}

\begin{table}[H]
\centering
\begin{tabular}{p{2.5cm} c c c}
\hline
\textbf{System}  & \textbf{Rouge-1} & \textbf{Rouge-2} & \textbf{Rouge-SU4} \\
\hline
Random & 0.302 & 0.046 & 0.088 \\
Lead & 0.312 & 0.058 & 0.102 \\
DSDR-non & 0.396 & 0.074 & - \\
MDS-Sparse+div       & 0.353 & 0.055 & 0.112 \\
MDS-Sparse-div       & 0.354 & 0.064 & 0.117 \\
Ours       & \textbf{0.403} & \textbf{0.092} & \textbf{0.146} \\
\hline
\end{tabular}
\caption{Results on DUC 2007.}
\label{tab:duc2007}
\end{table}



\subsection{Case Study}

Based on the news and comments of the topic ``Bitcoin exchange Mt. Gox goes offline'', we generate two summaries with our model considering comments (Ours) and ignoring comments (Ours-noC) respectively.
The summaries and ROUGE evaluation are given in Table \ref{tab:case1}. All the ROUGE values of our model considering comments are better than those ignoring comments with large gaps.
The sentences in \emph{\textbf{italic bold}} of the two summaries are different. By reviewing the comments of this topic, we find that many comments are talking about ``The company had lost 744,000 Bitcoins ...'' and ``Anonymity prevents reversal of transactions.'', which are well identified by our model.

\begin{small}
\begin{table}[H]
\centering
\begin{tabular}{|p{2.5cm} | c| c| c|}
\hline
\textbf{System}  & \textbf{Rouge-1} & \textbf{Rouge-2} & \textbf{Rouge-SU4} \\
\hline
Ours-noC & 0.365 & 0.097 & 0.126 \\
\hline
\multicolumn{4}{|p{8.1cm}|}{
\footnotesize
Mt. Gox went offline today , as trading on the Tokyo-based site came to a screeching halt.  \emph{\textbf{A withdrawal ban imposed at the exchange earlier this month. Deposits are insured by the government.}} The sudden closure of the Mt. Gox Bitcoin exchange sent the virtual currency to a three-month low on Monday
the currency's value has fallen to about \$470 from \$550 in the past few hours. The statement from the Bitcoin companies on Monday night , which was not signed by Mr. Silbert , are committed to the future of Bitcoin and the security of all customer funds.}\\
\hline
Ours       & \textbf{0.414} & \textbf{0.124} & \textbf{0.164}\\
\hline
\multicolumn{4}{|p{8.1cm}|}{
\footnotesize
Mt. Gox went offline today , as trading on the Tokyo-based site came to a screeching halt. \emph{\textbf{The company had lost 744,000 Bitcoins in a theft that had gone unnoticed for years.}} The sudden closure of the Mt. Gox Bitcoin exchange sent the virtual currency to a three-month low on Monday. The currency's value has fallen to about \$470 from \$550 in the past few hours.  \emph{\textbf{Anonymity prevents reversal of transactions.}} The statement from the Bitcoin companies on Monday night , which was not signed by Mr. Silbert , are committed to the future of Bitcoin and the security of all customer funds.}\\
\hline
\end{tabular}
\caption{Generated summaries for the topic ``Bitcoin exchange Mt. Gox goes offline''.}
\label{tab:case1}
\end{table}
\end{small}

We also present an entity rewriting case study. For person name ``Dong Nguyen'' in the topic ``Flappy Bird'', the summary without entity rewriting contains different mention forms such as  ``Dong Nguyen'', ``Dong'' and ``Nguyen''. After rewriting, ``Dong'' is replaced by ``Nguyen'', which makes the co-reference mentions clearer. As expected, there is only one full-form mention, such as ``Nguyen Ha Dong, a Hanoi-based game developer'' ``Shuhei Yoshida, president of Sony Computer Entertainment Worldwide Studios'', and ``The Australian Maritime Safety Authority's Rescue Coordination Centre, which is overseeing the rescue '', in each summary.

\section{Conclusion}
We propose a new MDS paradigm called reader-aware multi-document summarization (RA-MDS). To tackle this RA-MDS problem, we propose a sparse-coding-based method jointly considering news reports and reader comments.
We propose a compression-based unified optimization framework which explores a finer syntactic unit, namely, noun/verb phrase, to generate compressive summaries, and meanwhile it conducts entity rewriting
aiming at better linguistic quality.
In this work, we also generate a data set for RA-MDS task. 
The experimental results show that our framework
can achieve good performance and outperform state-of-the-art unsupervised systems.



\bibliographystyle{named}
\bibliography{ijcai15}

\end{document}